\documentclass[journal]{IEEEtran}

\usepackage{ntheorem}
\theoremheaderfont{\rmfamily\bfseries\upshape}
\theorembodyfont{\rmfamily\mdseries\itshape}
\theoremseparator{.}
\newtheorem{definition}{Definition}

\newtheorem{thm}{Theorem}

\usepackage{bbm}
\usepackage{upgreek}
\usepackage{amsmath}
\usepackage{xcolor}
\usepackage{graphicx}
\usepackage{booktabs}
\usepackage{amssymb}
\usepackage{textcomp}
\usepackage{amsmath,bm}
\usepackage{makecell}
\usepackage{multirow}
\usepackage{caption}
\usepackage{float}
\usepackage{colortbl}
\usepackage{mathrsfs}
\usepackage{changepage}
\usepackage[colorlinks]{hyperref}
\usepackage{color}
\usepackage{soul}
\usepackage{tcolorbox}
\usepackage[numbers,sort&compress]{natbib}
\makeatletter  
\newif\if@restonecol  
\makeatother

\usepackage[linesnumbered,ruled,vlined]{algorithm2e}
\usepackage{algpseudocode}  
\usepackage{amsmath}  

\usepackage{tikz,xcolor,hyperref}
\definecolor{lime}{HTML}{A6CE39}
\DeclareRobustCommand{\orcidicon}{%
    \begin{tikzpicture}
    \draw[lime, fill=lime] (0,0) 
    circle [radius=0.16] 
    node[white] {{\fontfamily{qag}\selectfont \tiny ID}};    \draw[white, fill=white] (-0.0625,0.095) 
    circle [radius=0.007];    \end{tikzpicture}
    \hspace{-2mm}}
\foreach \x in {A, ..., Z}{%
    \expandafter\xdef\csname orcid\x\endcsname{\noexpand\href{https://orcid.org/\csname orcidauthor\x\endcsname}{\noexpand\orcidicon}}
    }
\hypersetup{hidelinks}

\ifCLASSINFOpdf

\else

\fi

\begin{document}

\IEEEoverridecommandlockouts
\IEEEpubid{\begin{minipage}[t]{\textwidth}\ \\[1pt]
        \centering\footnotesize{\begin{tcolorbox}[left = 0.5mm, right = 0.5mm, top = 0.5mm, bottom = 0.5mm]\copyright This work has been submitted to the IEEE for possible publication. Copyright may be transferred without notice, after which this version may no longer be accessible.\end{tcolorbox}}
\end{minipage}}

\title{A Scalable Test Problem Generator for Sequential Transfer Optimization}

\author{Xiaoming Xue\orcidA{},
		Cuie Yang\orcidC{},~\IEEEmembership{~Member,~IEEE},
		Liang Feng\orcidD{},
		Kai Zhang\orcidF{}, \IEEEmembership{~Member,~IEEE}, \par
		Linqi Song\orcidG{},~\IEEEmembership{~Member,~IEEE},
		and Kay Chen Tan\orcidH{},~\IEEEmembership{~Fellow,~IEEE}
\thanks{Xiaoming Xue and Linqi Song are with the Department of Computer Science, City University of Hong Kong, Hong Kong SAR, China and also with the City University of Hong Kong Shenzhen Research Institute, Shenzhen 518057, China (e-mail: xminghsueh@gmail.com; linqi.song@cityu.edu.hk).}
\thanks{Cuie Yang is with the State Key Laboratory of Synthetical Automation for Process Industries, Northeastern University, Shenyang 110819, China (e-mail: cuieyang@outlook.com).}
\thanks{Liang Feng is with the College of Computer Science, Chongqing University, Chongqing 400044, China (e-mail: liangf@cqu.edu.cn).}
\thanks{Kai Zhang is with the Civil Engineering School, Qingdao University of Technology, Qingdao 266520, China (e-mail: zhangkai@qut.edu.cn).}
\thanks{Kay Chen Tan is with the Department of Computing, Hong Kong Polytechnic University, Hong Kong SAR, China (e-mail: kctan@polyu.edu.hk).}
}

\maketitle

\begin{abstract}
Sequential transfer optimization (STO), which aims to improve the optimization performance on a task of interest by exploiting the knowledge captured from several previously-solved optimization tasks stored in a database, has been gaining increasing research attention over the years. %
However, despite the remarkable advances in algorithm design, the development of a systematic benchmark suite for comprehensive comparisons of STO algorithms received far less attention. %
Existing test problems are either simply generated by assembling other benchmark functions or extended from specific practical problems with limited scalability. %
The relationships between the optimal solutions of the source and target tasks in these problems are also often manually configured, limiting their ability to model different similarity relationships presented in real-world problems. %
Consequently, the good performance achieved by an algorithm on these problems might be biased and hard to be generalized to other problems. %
In light of the above, in this study, we first introduce four concepts for characterizing STO problems and present an important problem feature, namely similarity distribution, which quantitatively delineates the relationship between the optima of the source and target tasks. %
Then, we present the general design guidelines of STO problems and a particular STO problem generator with good scalability. %
Specifically, the similarity distribution of a problem can be easily customized, enabling a continuous spectrum of representation of the diverse similarity relationships of real-world problems. %
Lastly, a benchmark suite with 12 STO problems featured by a variety of customized similarity relationships is developed using the proposed generator, which would serve as an arena for STO algorithms and provide more comprehensive evaluation results. %
The source code of the problem generator is available at \textcolor{magenta}{\url{https://github.com/XmingHsueh/STOP-G}}. %
\end{abstract}


\IEEEpeerreviewmaketitle


\section{Introduction}
Endowing optimizers with the ability to transfer knowledge from possibly related tasks, known as transfer optimization~\cite{gupta2017insights,lin2023multiobjective}, for better solving efficiency on tasks of interest has been receiving increasing research popularity over the years~\cite{tan2021evolutionary,wu2023diversified}, and has been successfully applied in many practical applications, such as software test~\cite{sagarna2016concurrently}, vehicle routing problems~\cite{feng2022towards}, production optimization~\cite{yao2021self}, well placement optimization~\cite{qi2023evolutionary}, and hyperparameter tuning of deep neural networks~\cite{chen2022scaling}. %
Sequential transfer optimization (STO)~\cite{gupta2017insights} is one of the representative paradigms in transfer optimization, in which a number of previously-solved tasks in a knowledge base serves as source tasks while the task at hand acts as target task. %
Existing STO algorithms can be roughly divided into two categories: 1) learning to optimize-based approaches that predict the optimum of the target task with a learned mapping between the features and the optima of the source tasks~\cite{shen2019lorm,li2021overview,bengio2021machine}, and 2) knowledge transfer-assisted search methods that improve the search performance of optimizers by transferring the knowledge extracted from the source tasks~\cite{cunningham1997case,feurer2015initializing,da2018curbing}. %
Over the decades, the latter has been gaining more and more research attention due to its compatibility with the state-of-the-art approaches in optimization and domain information-agnostic implementation~\cite{louis2004learning,feng2017autoencoding,shakeri2022scalable}. %

In knowledge transfer-assisted optimization, there are three main types of transferable objects: algorithm-related information~\cite{hutter2009paramils}, model-based features~\cite{hauschild2012using}, and task solutions~\cite{louis2004learning}. %
Specifically, algorithm configuration~\cite{hoos2011automated,huang2019survey}, algorithm selection~\cite{rice1976algorithm,smith2009cross} and algorithm portfolio~\cite{tang2021few} are three representative approaches using algorithm-based information~\cite{liu2017experience}. %
Model biasing \cite{wang2018regret,santana2012structural} and model aggregation~\cite{friess2020improving,min2019multiproblem} are two common approaches that transfer model-related features. %
For algorithms considering task solutions for knowledge transfer, kernelization of cross-task solutions~\cite{min2021generalizing,tighineanu2022transfer} and utilization of optimized solutions ~\cite{zhang2019multisource,xue2021evolutionary} are two popular approaches primarily studied in the literature. %
It is important to note that the three categories of transferable objects show a synergistic relationship instead of competing with one another, in view of the fact that they can be leveraged in a collaborated fashion towards better performance enhancement. %

In comparison with the proliferation of STO algorithms, the design of benchmark problems for STO received far less attention. %
Generally, there are two main motivations behind the benchmark design~\cite{huppler2009art}: 1) rapid, fair, repeatable and verifiable algorithm testings are achievable in synthetic environments and 2) top-notch algorithms with their promising results obtained on a benchmark suite are expected to aptly generalize to real optimization problems with relevant natures reflected by the benchmark. %
It is noted that such relevance for guiding the benchmark design should be supported by the foundation of the effectiveness of algorithms being developed. %
In view of the fact that the knowledge reasoning in transfer optimization is \emph{analogy}\footnote{As a reasoning method, analogy is opposed to deduction and induction.}, the similarity (also called relatedness) between tasks is supposed to the most pertinent feature needed to be adequately considered when designing benchmark problems. %
Thus, to enable a more comprehensive evaluation of different transfer algorithms, a set of problems featured by a broad spectrum of similarity relationships should be developed~\cite{scott2023first}. %
At present, there are two approaches widely used for generating synthetic STO problems (STOPs), which consist of both source and target optimization tasks: %

\emph{1) Synthesis with elemental functions}: Synthesizing STOPs with elemental functions is a common way of generating synthetic test problems in transfer optimization~\cite{feng2017autoencoding,da2018curbing,zhang2019multisource}. %
The elemental functions in single-objective optimization~\cite{friess2020improving} and those from several popular benchmark suites in multiobjective optimization are commonly used for this purpose, including ZDT family~\cite{zitzler2000comparison}, DTLZ family~\cite{deb2005scalable}, and WFG family~\cite{huband2006review}. %
Without loss of generality, the optima of most of these functions are set to be similar vectors as they will be solved separately. %
Notably, this trivial setting becomes consequential in transfer optimization: the transfer of elite solutions across these tasks always yield significant convergence speedups. %
However, such high task similarity in terms of optimum is quite rare, making the promising results obtained on such problems hard to be guaranteed in practice. %
To simulate STOPs that exhibit task variations with respect to optimum, several studies use the parameterization strategy to manipulate the optimal solutions of source tasks~\cite{min2019multiproblem,xue2021evolutionary,yu2023experience}. %
However, the relationship between the optima of the source and target tasks is not well analyzed, although the performance of most STO algorithms is susceptible to such relationship. %
Besides, the relationship produced by a specific parameterization scheme can only represent the relationship of a particular type of STOPs, making the algorithm tests highly problem-specific. %

\begin{figure}[ht]
	\centering
	\includegraphics[width=3.4in]{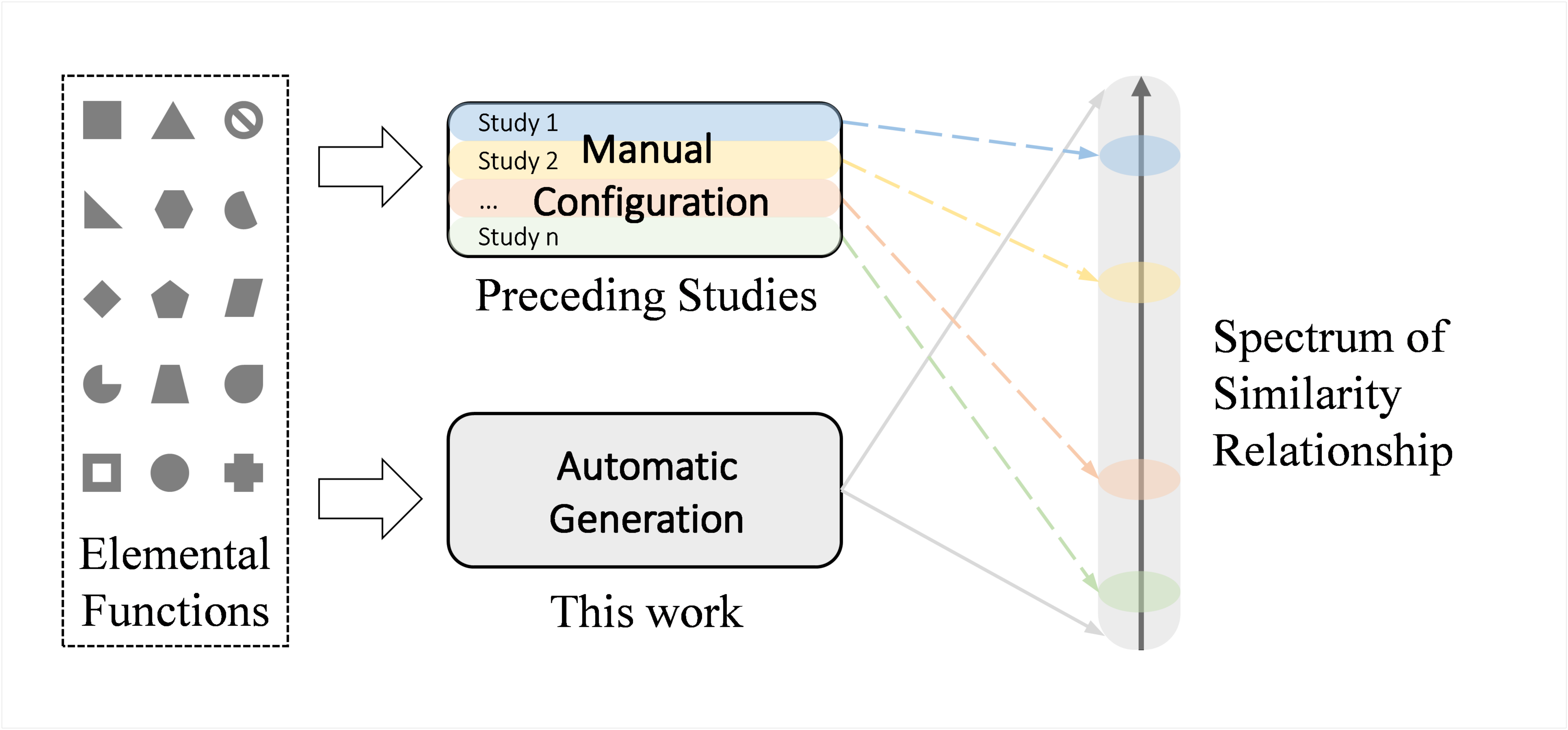}
	\caption{An illustrative comparison between the test problems in the preceding studies and this work with respect to their capabilities of representing the similarity relationship.}
	\label{fig:spectrum}
\end{figure}

\emph{2) Extension of practical problems}: The other way of generating STOPs is to perturb the parameters or features of a practical problem for generating its source tasks~\cite{louis2004learning,da2018curbing,zhang2019multisource}. %
However, these STOPs still suffer from the aforementioned issues. %
For example, the source tasks are generated according to the parameters of the target task to ensure a certain degree of similarity in~\cite{louis2004learning}. %
Similarly, the source tasks are synthesized by sampling the parameters of the target task with respect to objective or constraint in \cite{zhang2019multisource}. %
However, the prespecified high similarity between the tasks in these STOPs is very rare in real-world problems. %
In summary, the optima of the source and target tasks in the existing synthetic STOPs are always manually configured, making the resultant similarity relationships irregularly yet discretely distributed, as illustrated in Fig. \ref{fig:spectrum}. %
The similarity relationships of the problems in an individual study only represent a particular type of relationships in the spectrum, on which one often achieves promising results without deeper insights about how their algorithms perform when the relationship varies~\cite{scott2023first}. %
Therefore, it is desirable to develop a problem generator with configurable similarity relationship, enabling a continuous spectrum of representation of the diverse similarity relationships presented in real-world STOPs. %
With the generator, one can produce a set of STOPs with diverse similarity relationships to evaluate different STO algorithms, enabling a more comprehensive investigation of the algorithms' performance. %
More importantly, such a set of test problems can serve as a benchmark suite for rapid testing of different STO algorithms in synthetic environments. %

With the above in mind, this paper develops a scalable test problem generator of STOPs. %
Firstly, we present an important problem feature, namely similarity distribution, to quantitatively delineate the similarity relationship between the optima of the source and target tasks in an STOP. %
This problem feature is founded on a few newly defined basic concepts that characterize STOPs. %
Subsequently, we present a set of design guidelines, along with an inverse transform sampling-based strategy, enabling us to generate any STOP with the desired similarity relationship that can be explicitly specified in advance. %
Notably, the general representation capability of the proposed generator that can produce STOPs with a continuous spectrum of representation of the similarity relationship, as demonstrated in Fig. \ref{fig:spectrum}, is theoretically analyzed. %
On the whole, the proposed generator has three commendable attributes: %
\begin{itemize}
\item General representation of the similarity relationship: a spectrum of diverse similarity relationships between the optima of the source and target tasks of real-world STOPs can be achieved. %
\item Optimizer-independency: the generator does not rely on specific optimizers, making it coherent with the state-of-the-art approaches in optimization. %
\item High scalability: high-dimensionality, single-source transfer, objective conflicts can be configured easily by modifying the parameters of the problem generator. %
\end{itemize}
Lastly, a benchmark suite with 12 STOPs featured by three types of similarity relationships, i.e., high similarity, mixed similarity, and low similarity, is developed using the generator. %

The remaining of this paper is structured as follows. %
In Section II, we will present a few rudimentary concepts to introduce a significant and problem-dependent feature of STOPs, namely similarity distribution, which will be elaborated through an illustrative example. %
Subsequently, Section III furnishes the detailed design principles and a problem generator for producing STOPs with customizable similarity distribution. %
In Section IV, a benchmark suite with 12 individual STOPs is developed using the problem generator. %
Lastly, Section V concludes the paper. %


\section{Basic Concepts}

\begin{figure}[ht]
	\centering
	\includegraphics[width=3.4in]{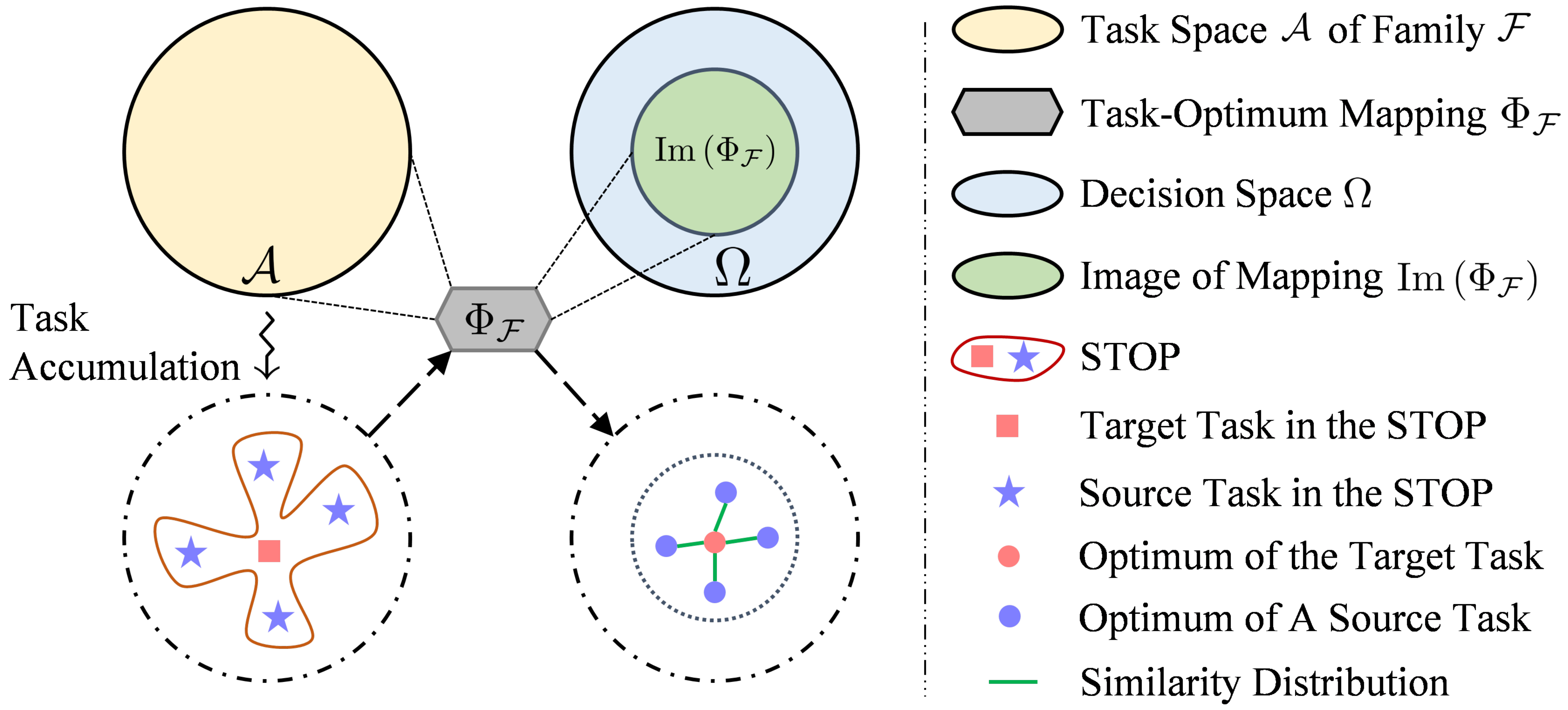}
	\caption{An illustration of the newly defined concepts.}
	\label{fig:concepts}
\end{figure}

In this section, we first present a set of fundamental concepts whose interrelationship with the well-established notions in transfer optimization are illustrated in Fig. \ref{fig:concepts}. In view of the multiple individual tasks (i.e., the source and target tasks) in an STOP, we introduce a concept, known as \emph{task family}, to refer to a task set from which the source and target tasks are derived. Furthermore, given that many STO algorithms transfer knowledge in the form of elite solution(s), it is imperative to scrutinize the relationship between the optimal solutions of the source and target tasks. Thus, we introduce another concept, namely \emph{task-optimum mapping}, to link a particular task to its optimum. Consequently, for a task family, a set of optima corresponding to the elementary tasks exist, which is denoted as the image of the task-optimum mapping. Based on this, we further define \emph{optimum coverage} to describe the relative size of the image over the decision space. Finally, by measuring the similarity between two tasks based on the distance of their optimal solutions, we define \emph{similarity distribution} to characterize the optimum-based similarity relationship between the $k$ source tasks and the target task of an STOP. To aid readers in assimilating the newly defined concepts, we provide an illustrative example of these concepts at the end of this section.

\subsection{Task Family}

Firstly, we present the concept of task family that refers to a set of tasks serving as the origin of the tasks in an STOP,
\begin{definition} A task family $\mathcal{F}\left(\mathbf{x};\mathcal{A}\right)$ consists of a set of elementary optimization tasks $f\left(\mathbf{x};\mathbf{a}\right)$ with the decision variable $\mathbf{x}$ in a decision space $\Omega$, which is given by:
\begin{equation}
\mathcal{F}\left(\mathbf{x};\mathcal{A}\right) = \{f\left(\mathbf{x};\mathbf{a}\right)|\mathbf{a}\in\mathcal{A}\},\,\,\,\mathbf{x}\in\Omega
\end{equation}
where $\mathcal{A}$ is a task space that contains all the realizations of the elementary task, $\mathbf{a}$ denotes a feature vector that characterizes a particular elementary optimization task in $\mathcal{F}$.
\end{definition}

The notion of a task family can be aptly exemplified through various optimization problems. For instance, the minimization of $n$-dimensional quadratic functions featuring different coefficient vectors can be deemed a task family. As such, a specific realization of the coefficient vector deterministically defines the particular quadratic optimization task. Furthermore, in combinational optimization~\cite{wang2023multiobjective}, traveling salesman optimization tasks with $n$ cities can also be deemed a task family. The coordinate information can characterize a specific task. Notably, the task family is a fairly general concept, with its elementary tasks contingent on real-world applications of interest. Given a task family $\mathcal{F}$, we can further postulate that its elementary task $f\in\mathcal{F}$ is a random variable with a latent distribution, which is given by:
\begin{equation}
f\left(\mathbf{x};\mathbf{a}\right)\sim f\left(\mathbf{x};p_{\mathcal{A}}\right)
\label{eq:latent_distribution}
\end{equation}
where $p_{\mathcal{A}}$ denotes the distribution of the parametric features. The feature vector of a task family is also defined as operating conditions belonging to an auxiliary space in~\cite{gupta2017insights}.

According to the definition of the task family, knowledge transfer can be divided into two categories: \emph{intra-family transfer} and \emph{inter-family transfer}. Specifically, in the intra-family transfer, the source and target tasks in an STOP are drawn from the same family, thus sharing similar landscape modalities. By contrast, the source and target tasks of an STOP in the inter-family transfer are from different task families.

\subsection{Task-Optimum Mapping}
Given a task family $\mathcal{F}\left(\mathbf{x};\mathcal{A}\right)$, its elementary task $f\left(\mathbf{x};\mathbf{a}\right)\in\mathcal{F}$ is a regular optimization problem. Without loss of generality, the elementary tasks are considered to be single-objective minimization optimization tasks\footnote{A maximization task can be converted to a minimization task via multiplying its objective function by -1.}, as given by,
\begin{equation}
\mathbf{x}^* = \min_{\mathbf{x}\in\Omega} f\left(\mathbf{x};\mathbf{a}\right)
\end{equation}
where $\mathbf{x}^*$ denotes the optimal solution. Generally, for a task with known features, its optimum exists deterministically and uniquely\footnote{Optimization tasks with multiple global optima are not considered herein.}. Thus, we define task-optimum mapping to link the features of a task to its optimum.
\begin{definition} Given a task family $\mathcal{F}\left(\mathbf{x};\mathcal{A}\right)$, the task-optimum mapping maps the features of any task $f\left(\mathbf{x};\mathbf{a}\right)$ to its optimum:
\begin{equation}
\Phi_{\mathcal{F}}\left(\mathbf{a}\right) = \mathbf{x}^*,\,\,\,\mathbf{x}^*\subseteq\Omega
\end{equation}
where $\Phi_{\mathcal{F}}:\mathcal{A}\to\Omega$ is the task-optimum mapping of $\mathcal{F}$.
\end{definition}

The task-optimum mapping defined herein is a fairly general concept that can refer to the optimization process of any optimization task of interest. 
For tasks with an explicit relation between features and optimum, an analytical solution of the task-optimum mapping is often available. 
For example, the optimal solution of a convex quadratic optimization task can be analytically derived based on its coefficients~\cite{boyd2004convex}. 
However, it is more common for tasks to have an implicit relationship between features and optimum. 
In such cases, an optimizer is typically utilized to search for the optimum through iterative evaluations of the objective function. Notably, the concept of task-optimum mapping still holds despite the implicity, thereby allowing us to learn it from optimization data. Specifically, given a set of previously-solved tasks $\{f\left(\mathbf{x};\mathbf{a}_s^i\right),i=1,...,k\}$ from a task family $\mathcal{F}\left(\mathbf{x};\mathcal{A}\right)$, their optima are represented as $\{\mathbf{x}_{si}^{*},i=1,...,k\}$. The task-optimum mapping of this family can be learned by a supervised learning algorithm:
\begin{equation}
\widetilde{\Phi}_{\mathcal{F}}=\min_{\Phi\in\Theta}\sum_{i=1}^{k}l\left(\Phi\left(\mathbf{a}_s^i\right),\mathbf{x}_{si}^{*}\right)
\end{equation}
where $\Theta$ is a mapping space with candidate mappings, $l\left(\cdot,\cdot\right)$ denotes a metric function for measuring the discrepancy between the predicted and true optima. Given an unsolved target task $f\left(\mathbf{x};\mathbf{a}_t\right)\in\mathcal{F}$, its optimum can be predicted as follows:
\begin{equation}
\widetilde{\mathbf{x}}_{t}^{*} = \widetilde{\Phi}_{\mathcal{F}}\left(\mathbf{a}_t\right)
\end{equation}

Moreover, for particular task families whose task-optimum mappings satisfy the uniform continuity, we have $\forall x,y\in\mathcal{A}$, $\exists\delta>0$, $\forall\epsilon>0$:
\begin{equation}
|x-y|<\delta\Rightarrow|\Phi\left(x\right)-\Phi\left(y\right)|<\epsilon
\end{equation}

This property suggests that two similar tasks in the feature space possess optimal solutions in close proximity, which forms the foundation for many transfer algorithms that use problem features to measure the similarity. A few representative methods can be found in \cite{yin2002genetic,burke2006case,yamamoto2015refined,oman2001using}. For a task family, the image of its task-optimum mapping is the set of optima associated with all the elementary tasks in the decision space~\cite{deisenroth2020mathematics}, as illustrated in Fig. \ref{fig:concepts}. The following subsection introduces a concept named optimum coverage, which is used for quantifying the relative size of this image.

\subsection{Optimum Coverage}
Given a task family $\mathcal{F}\left(\mathbf{x};\mathcal{A}\right)$, its task-optimum mapping $\Phi_{\mathcal{F}}$ often has the following two properties:
\begin{itemize}
\item[--] \emph{Non-injective}: $\exists \mathbf{a}_i,\mathbf{a}_j\in\mathcal{A}:\Phi\left(\mathbf{a}_i\right)=\Phi\left(\mathbf{a}_j\right)\nRightarrow \mathbf{a}_i=\mathbf{a}_j$.
\item[--] \emph{Non-surjective}: $\Phi\left(\mathcal{A}\right)\neq\Omega$. 
\end{itemize}

The first property implies that different tasks from the same family may share a common optimum. For instance, a minor change of the edge information in a short-path problem may not result in the alteration of the shortest path. The second property means that the image of the task-optimum mapping, i.e., $\mathrm{Im}\left(\Phi_{\mathcal{F}}\right)$, is distributed over a subregion of the decision space. Fig. \ref{fig:concepts} shows the relationship between the task space $\mathcal{A}$ and the decision space $\Omega$ connected by the task-optimum mapping $\Phi_{\mathcal{F}}$. The image of $\Phi_{\mathcal{F}}$ is the set of optimal solutions corresponding to the elementary tasks in $\mathcal{A}$. According to Eq. \eqref{eq:latent_distribution} and the definition of the task-optimum mapping, we have the distribution of the optimal solutions as follows:
\begin{equation}
\Phi_{\mathcal{F}}\left(\mathbf{a}\sim p_{\mathcal{A}}\right) = \mathbf{x}^* \sim q_{\mathcal{A}}\left(\mathbf{x}^*\right)
\label{eq:opt_distribution}
\end{equation}
where $q_{\mathcal{A}}$ is the optimum distribution of the elementary tasks in $\mathcal{F}$. Here, the event space of $\mathbf{x}^*$ is equivalent to the image of $\Phi_{\mathcal{F}}$. Next, we define optimum coverage to describe the relative size of such image over the decision space.

\begin{definition} Given a set of families $\Theta=\{\mathcal{F}_1,...,\mathcal{F}_m\}$ from which the source and target tasks of an STOP are derived, optimum coverage is defined as the ratio of the image $\mathrm{Im}\left(\Phi_{\Theta}\right)$ to the common space $\Omega_c$:
\begin{equation}
\gamma=
\begin{cases}
\int_{x\in\mathrm{Im}\left(\Phi_{\Theta}\right)}\mathrm{d}x/\int_{x\in\Omega_c}\mathrm{d}x,&\mathrm{continuous}\,\mathrm{tasks}\\
\sum_{x\in\mathrm{Im}\left(\Phi_{\Theta}\right)}1/\sum_{x\in\Omega_c}1,&\mathrm{discrete}\,\mathrm{tasks}
\end{cases}
\label{eq:coverage_rate}
\end{equation}
where $\gamma$ is the optimum coverage, $m$ is the number of task families. $\mathrm{Im}\left(\Phi_{\Theta}\right)$ is $\mathrm{Im}\left(\Phi_{\mathcal{F}}\right)$ for a single family, while it equals to $\mathrm{Im}\left(\Phi_{\mathcal{F}_1}\right)\cup...\cup\mathrm{Im}\left(\Phi_{\mathcal{F}_m}\right)$ for multiple families. Apparently, $m$ is equal to 1 in the intra-family transfer while it is greater than 1 in the inter-family transfer.
\end{definition}

In particular, $\gamma\approx0$ if all the tasks in a family share a common optimum, $\gamma=1$ when the optimal solutions spread over the whole search space. It is noted that most practical problems fall between these two extreme cases. The optimum coverage of a task family can, to some extent, signify one's prior knowledge pertaining to this particular family. Usually, while dealing with a task family where one has little prior knowledge available, it is common to adopt a large decision space owing to one's meager perception regarding the optimal solutions. In such situations, the optimal coverage is small. Conversely, if certain knowledge regarding the optimal solutions is available, one prefers choosing a narrow decision space resulting in a large optimum coverage. Thus, the optimum coverage is a highly problem-dependent property of the task family. Furthermore, to characterize the relationship between the optima of the source and target tasks in an STOP, we define a concept named similarity distribution in what follows.

\subsection{Similarity Distribution}
When the source and target tasks of an STOP are derived from a family set $\Theta=\{\mathcal{F}_1,...,\mathcal{F}_m\}$, the accumulation of the tasks can be seen as an increasing number of independent samples drawn from the joint task distribution of the $m$ task families. 
Suppose the source and target tasks are denoted by $\{T_{s1},T_{s2},...,T_{sk}\}$ and $T_t$, one can quantify the relationship between the $i$th source task and the target task as follows:
\begin{equation}
R_i = Q\left(T_{si},T_{t}\right),\,\,\,i=1,2,...,k
\label{eq:relationship}
\end{equation}
where $R_i$ denotes the relationship value between the $i$th source task and the target task, $Q\left(\cdot\right)$ is a relationship measurement function. Apparently, $\mathbf{R}=\{R_{1},R_{2},...,R_{3}\}$ depicts the overall relationship between the $k$ source tasks and the target task.

According to Eq. \eqref{eq:relationship}, we can see that the meaning of the relationship depends on two aspects, i.e., the relationship measurement function and the task information for calculating the relationship. 
For the measurement function, its choice mainly depends on the understanding of what kind of relationship has a significant impact on the performance of transfer algorithms, which could be from either prior beliefs or empirical evidence. 
For the task information, both problem-specific features (e.g., $\mathbf{a}$) or meta-features (e.g., the landscape smoothness) are valid for calculating the relationship\footnote{For black-box STOPs, only meta-features are available for the relationship measurement.}. 
For most transfer algorithms, their performance is susceptible to the similarity between the optimal solutions of the source and target tasks, especially those that transfer elite solutions~\cite{feng2017autoencoding,zhang2019multisource} or model information based upon solution data~\cite{da2018curbing,shakeri2022scalable}. 
Therefore, we propose to quantify the relationship using the optimum-based similarity in this work, which is given by,
\begin{figure}[ht]
	\centering
	\includegraphics[width=2.54in]{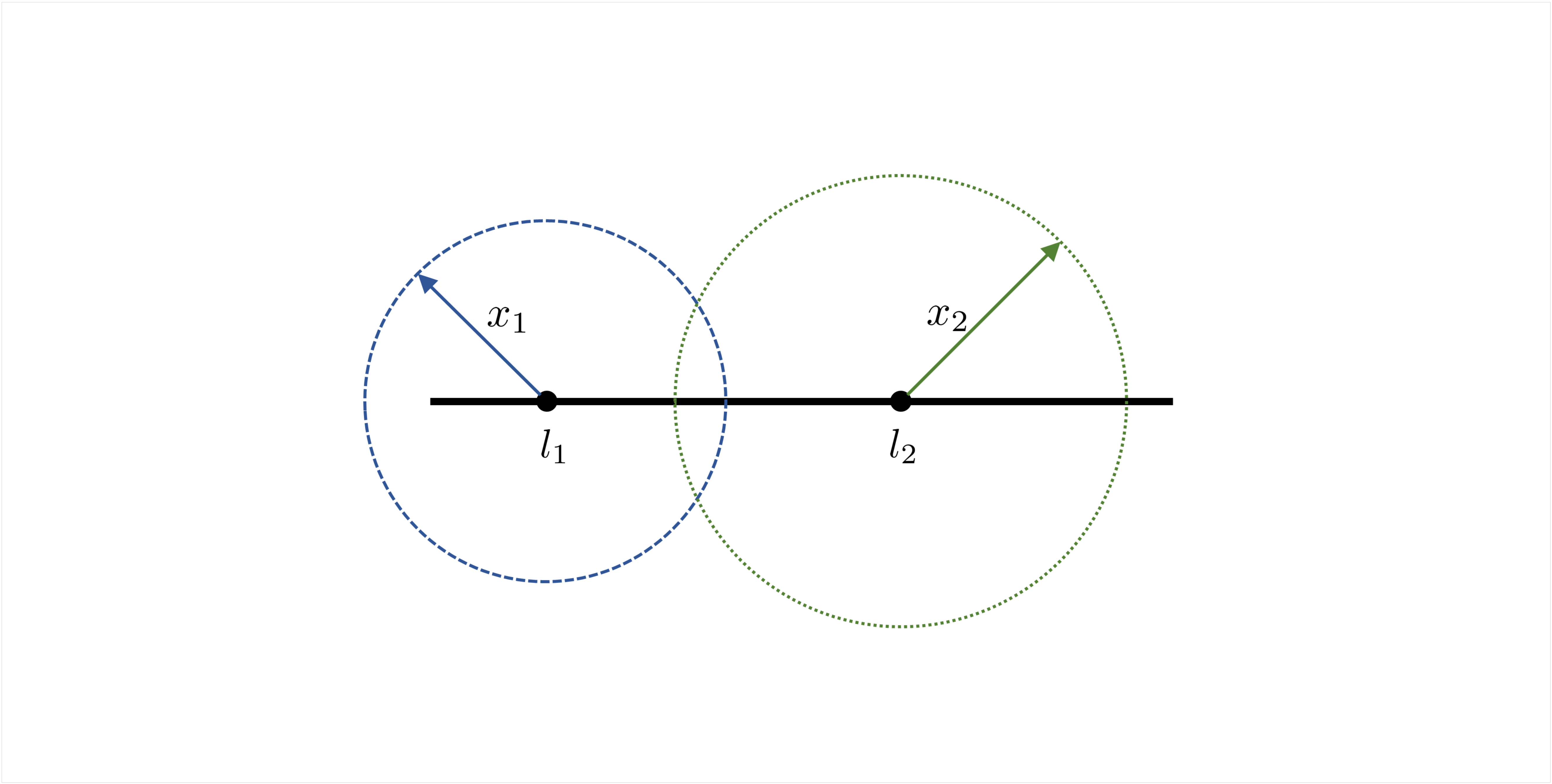}
	\caption{A schematic diagram of the interval coverage task.}
	\label{fig:example}
\end{figure}
\begin{equation}
R^s_i = 1 - D_{Cheb}\left(\mathbf{o}_{si},\mathbf{o}_{t}\right)=1-\max_j\left(|o_{si}^{j}-o_{t}^{j}|\right)
\label{eq:sim_relationship}
\end{equation}
where $R^s_i\in\left[0,\,1\right]$ denotes the similarity relationship value between the $i$th source task and the target task, $D_{Cheb}\left(\cdot,\cdot\right)$ is the Chebyshev distance function\footnote{Without loss of generality, other distance or correlation functions can also be used.}, $\mathbf{o}_{t}$ is the optimum of the target task in the common space, $\mathbf{o}_{si}$ denotes the optimum of the $i$th source task in the common space, $o_{t}^{j}$ and $o_{si}^{j}$ represent the $j$th variables of $\mathbf{o}_{t}$ and $\mathbf{o}_{si}$, respectively.

With Eq. \eqref{eq:sim_relationship}, we can quantify the relationship between the $k$ source tasks and the target task with respect to the optimum-based similarity using the set of similarity values $\mathbf{R^s}=\{R^s_{1},R^s_{2},...,R^s_{k}\}$. 
To eliminate the effect of the possibly varying number of source tasks in different STOPs, one can estimate a probability distribution function for $\mathbf{R^s}$ using any density estimation method, e.g., Parzen windows~\cite{bowman1997applied}. 
Without loss of generality, we employ the well-known density estimation method, i.e., the rescaled histogram, to estimate the probability density function for $\mathbf{R^s}$, which is given by
\begin{equation}
\begin{split}
g\left(s;\mathbf{R}^s\right) &= \frac{1}{k}\sum_{i=1}^{k}I_{(\frac{\lfloor sn\rfloor}{n},\frac{\lceil sn\rceil}{n}]}\left(R^s_i\right)\\
&=\frac{1}{k}\bigg| i \,|\,\frac{\lfloor sn\rfloor}{n}<R^s_i\le\frac{\lceil sn\rceil}{n},\,1\le i\le k\bigg|
\end{split}
\label{eq:histogram_estimate}
\end{equation}
where $g\left(s;\mathbf{R}^s\right)$ denotes the estimated density function of $\mathbf{R}^s$, $n$ is the number of equal-width bins of the histogram, $I_{(a,b]}\left(c\right)$ denotes the indicator function, which returns 1 if $c\in(a,b]$ and 0 otherwise, $\lceil sn\rceil=\lfloor sn\rfloor+1$ when $\lfloor sn\rfloor\in\mathbb{Z}$. %
It is important to note that there is no optimal number of bins as different bin sizes always reveal different features of the data. %
Generally, using wider bins where the density of $\mathbf{R}^s$ is low can reduce noise especially in the case of small $k$, while using narrower bins where the density is high yields greater precision to the density estimation. %

Throughout this paper, we refer to the estimated density function in Eq. \eqref{eq:histogram_estimate} as similarity distribution for brevity. %
It can be seen that the similarity distribution quantitatively characterizes the similarity relationship between the optimal solutions of the source and target tasks in an STOP, which is mainly affected by the following three factors: %
\begin{itemize}
\item Optimum coverage of the family set(s) of interest.
\item Optimum distribution induced by the task distribution.
\item Sampling randomness of the tasks in an STOP.
\end{itemize}

Next, we present a toy example to help readers digest all the defined concepts and demonstrate how the similarity distribution changes with the above three factors.

\begin{figure}[ht]
	\centering
	\includegraphics[width=3.4in]{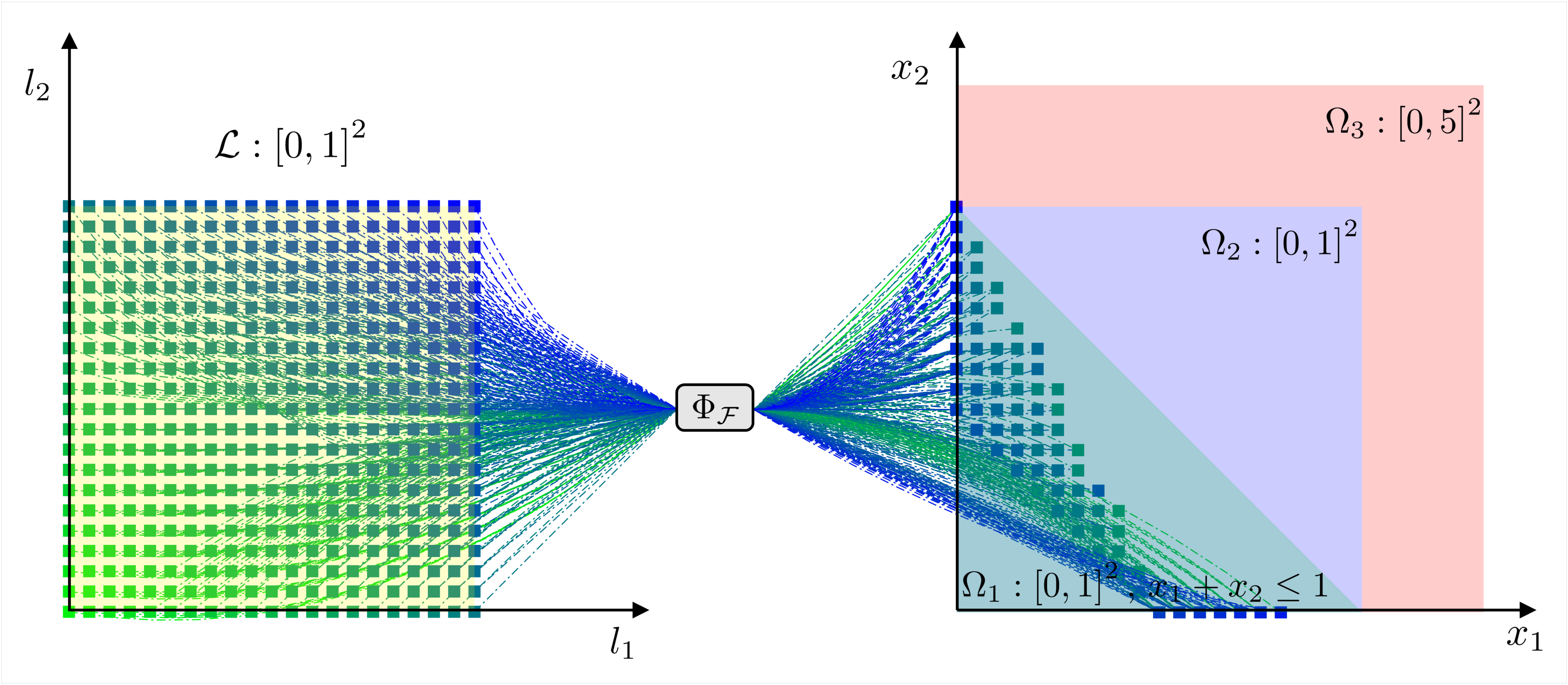}
	\caption{A large number of interval coverage tasks in $\mathcal{F}\left(\mathbf{x};\mathcal{L}\right)$ and the locations of their optima in $\Omega_1$, $\Omega_2$ and $\Omega_3$.}
	\label{fig:example_mapping}
\end{figure}

\subsection{A Toy Example}

Given an 1-dimensional interval and two stations within the interval, one aims to minimize the radiuses of the two stations while ensuring the coverage over the interval. %
Fig. \ref{fig:example} shows the schematic diagram of this optimization task. %
More formally, an interval coverage optimization task $f\left(\mathbf{x};\boldsymbol{l}\right)$ with the feature vector $\boldsymbol{l}=\left[l_1,l_2\right]$ (i.e., the locations of the two stations) can be formulated as follows: %
\begin{gather}
\min_{\mathbf{x}\in\Omega} x_1+x_2\\
\emph{s.t.\,\,}\mathrm{The\,\,interval\,\,is\,\,covered.} \notag
\end{gather}
where $\mathbf{x}=\left[x_1,x_2\right]$ denotes the decision vector that represents the radiuses of the two stations, $\Omega$ represents the decision space configured by a practitioner. %
It is noteworthy that all the box-constrained spaces from $\left[0,1\right]^2$ to $\left[0,\infty\right]^2$ are feasible settings with no influence on the optimal solutions. %
Next, we shall use this example to elaborate the concepts defined earlier and show the necessity of introducing them into STO.

\textbf{Task family}: The task family $\mathcal{F}\left(\mathbf{x};\mathcal{L}\right)$ contains an infinite number of elementary tasks characterized by the feature vector $\boldsymbol{l}$ in the feature space $\mathcal{L}:\left[0,1\right]^2$. %
An elementary task $f\left(\mathbf{x};\boldsymbol{l}\right)$ can be drawn from $\mathcal{F}\left(\mathbf{x};\mathcal{L}\right)$. %
To illustrate, we sample a large number of elementary tasks that are uniformly distributed over the feature space, as depicted on the left side of Fig. \ref{fig:example_mapping}. %
Each solid square within the shaded area (i.e., $\mathcal{L}$) represents a particular elementary interval coverage optimization task. 

For practitioners who conduct frequent re-optimization of such interval coverage tasks, it is the case of STO where a few previously-solved source tasks can be used for better solving a target task. %
That is why we introduce the task family to refer to the task set from which these tasks are derived.

\textbf{Task-optimum mapping}: The task-optimum mapping $\Phi_{\mathcal{F}}$ of this family maps the features of an interval coverage task to its optimum. %
To illustrate, we optimize all the elementary tasks sampled before and show the locations of their optima as the solid squares on the right side of Fig. \ref{fig:example_mapping}. %
The task-optimum mapping is displayed in the form of many dotted lines.

As mentioned earlier, we are interested in the similarity relationship between the optimal solutions of the source and target tasks in an STOP. %
Thus, we can use the task-optimum mapping to map a collection of tasks in the family to their optimal solutions for obtaining such relationship. %
The set of the optimal solutions of all the elementary tasks is known as the image of the task-optimum mapping, i.e., $\mathrm{Im}\left(\Phi_{\mathcal{F}}\right)$.

\begin{figure}[ht]
	\centering
	\includegraphics[width=3.4in]{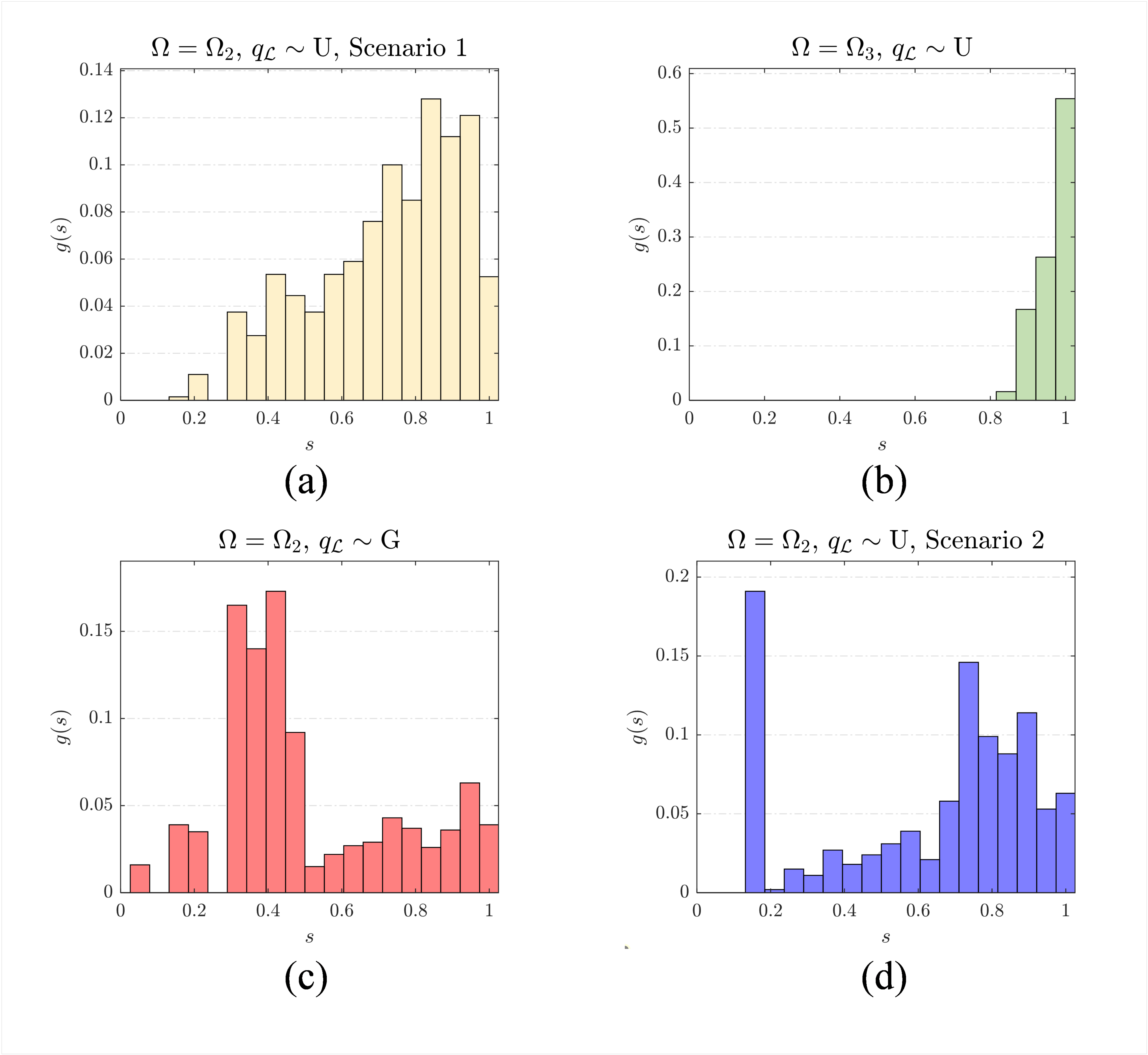}
	\caption{Similarity distributions of four interval coverage problems: (a) $\Omega=\Omega_2,\,q_{\mathcal{L}}\sim\mathrm{U},\,\mathrm{Scenario}\,1$; (b) $\Omega=\Omega_3,\,q_{\mathcal{L}}\sim\mathrm{U}$; (c) $\Omega=\Omega_2,\,q_{\mathcal{L}}\sim\mathrm{G}$; (d) $\Omega=\Omega_2,\,q_{\mathcal{L}}\sim\mathrm{U},\,\mathrm{Scenario}\,2$.}
	\label{fig:sds_com}
\end{figure}

\textbf{Optimum coverage}: The optimum coverage is defined as the ratio of the image $\mathrm{Im}\left(\Phi_{\mathcal{F}}\right)$ to the decision space. %
It is noted that the size of the decision space relies on our prior beliefs about the optimal solutions of the task family. %
If some priors are available, one always prefers to adopt a smaller decision space to make the optimizer find the optimum more efficiently. %
In this example, we consider three cases of decision spaces (i.e., $\Omega_1$, $\Omega_2$ and $\Omega_3$ in Fig. \ref{fig:example_mapping}) and calculate the optimum coverages by discretizing the decision spaces. %
The optimum coverages corresponding to the three decision spaces are $\gamma_1=0.34$, $\gamma_2=0.17$, and $\gamma_3=0.01$, respectively. %
It can be seen that a larger decision space leads to a lower optimum coverage.

As mentioned earlier, the optimum coverage is a problem-dependent feature and is governed by two factors: the intrinsic nature of a task family that determines the optimum distribution and human priors that influence the size of the decision space. %
Since the source and target tasks in an STOP are alway a subset of the elementary tasks in a task family (or multiple families in the inter-family transfer), their similarity relationship is sensitive to the optimum coverage, which will be examined further in what follows. %

\textbf{Similarity distribution}: In this example, we can construct an STOP by sampling an arbitrary number of elementary tasks from $\mathcal{F}\left(\mathbf{x};\mathcal{L}\right)$. %
Specifically, $10^3$ tasks are optimized and stored into the database to serve as the source tasks, while an unsolved task acts as the target task. %
The similarity relationship between the $10^3$ source tasks and the target task in terms of optimum can be described by the similarity distribution defined in Eq. \eqref{eq:histogram_estimate}, where the number of equal-width bins are set to 20. %
Next, we shall use the variable-controlling method to scrutinize the three aforementioned factors that are responsible for the variation of the similarity distribution in different STOPs.

Firstly, we shall investigate how the similarity distribution changes with the optimum coverage. %
The similarity distributions of two STOPs with the same optimum distribution $U\left[0,1\right]^2$ but different optimum coverages are shown in Fig. \ref{fig:sds_com}(a) and Fig. \ref{fig:sds_com}(b). %
It can be seen that the similarity distribution is highly susceptible to the optimum coverage. %
As the optimum coverage decreases (i.e., from $\gamma_2=0.17$ to $\gamma_3=0.01$), the overall similarity of the source tasks to the target task improve a lot. %
This is because the optimal solutions of the elementary tasks are closer in the common space when the optimum coverage is smaller, thus leading to the higher similarity values. %
Secondly, we investigate the influence of the optimum distribution on the similarity distribution. %
Two STOPs with the same optimum coverage of $\gamma_2=0.17$ but different optimum distributions are considered, whose similarity distributions are shown in Fig. \ref{fig:sds_com}(a) and Fig. \ref{fig:sds_com}(c). %
The optimum distributions are the uniform and Gaussian distributions. %
It can be seen that the two STOPs with distinct optimum distributions have entirely different similarity distributions. %
Lastly, we find that the sampling randomness of the source and target tasks also impacts the similarity distribution. %
As shown in Fig. \ref{fig:sds_com}(a) and Fig. \ref{fig:sds_com}(d), the similarity distributions of STOPs with the same optimum coverage and optimum distribution can differ a lot. %

According to the above analyses, it can be concluded that the similarity distribution is a highly problem-dependent feature. %
It is observed that even if two STOPs are from the same task family, their similarity distributions may be entirely different. %
However, the patterns of the similarity distribution in the existing test problems are considerably narrow, making them inadequate for the assessment of various STO algorithms. %
Thus, it is imperative to take this crucial feature into account when designing a problem generator of STOPs, to endow generated test problems with a broad spectrum of similarity relationship that better mimic real-world problems.


\section{Proposed Problem Generator}

In this section, we develop a problem generator of STOPs with configurable similarity distribution. %
Firstly, we propose a few design guidelines to show how to make the similarity distribution of an STOP configurable.
Its general representation capability is also theoretically analyzed and discussed. %
After that, a problem generator with detailed algorithmic implementation is presented. %
Lastly, we compare the important features of the proposed test problems with those in existing studies.

\subsection{Design Principles}
In a problem domain, one always sequentially encounters a number of optimization tasks with possibly different parameters under the same formulation, which can be deemed a task family defined in this work. %
When a certain number of intra-family source tasks are available, we can choose to trigger the intra-family transfer. %
However, intra-family source tasks may be scarce, especially at the early stage of solving the tasks from a family. %
In this case, one can employ some out-of-family tasks and let them serve as the source tasks, which is known as the inter-family transfer.

\begin{figure}[ht]
	\centering
	\includegraphics[width=3.4in]{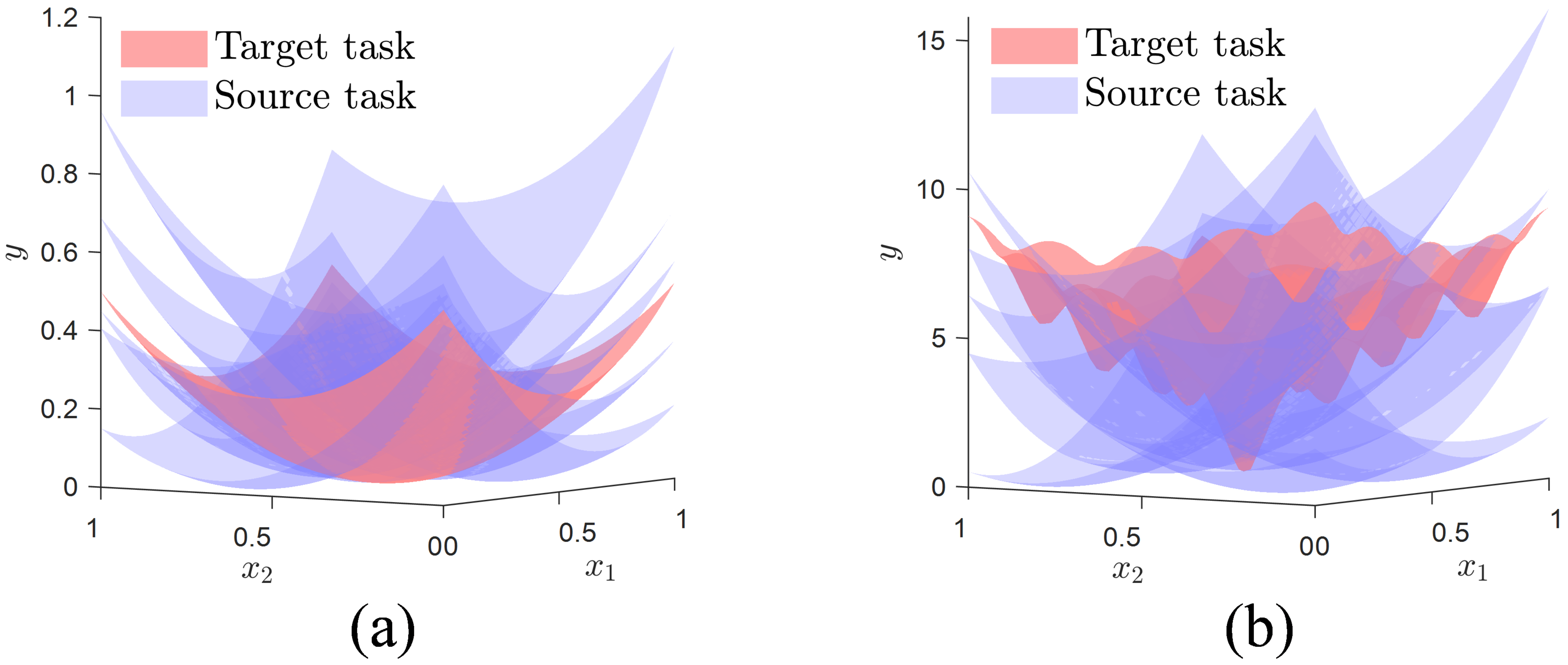}
	\caption{Examples of the intra-family and inter-family transfers: (a) the intra-family transfer; (b) the inter-family transfer.}
	\label{fig:intrainter}
\end{figure}

\subsubsection{Intra-Family Transfer \& Inter-Family Transfer} Given a task family, different elementary tasks in the family normally have distinct optimization landscapes and thus may have varying optimal solutions. %
Since the optimum is just one of the features of the optimization landscape, the landscape modality is more resistant to the change of tasks than the optimum. %
In other words, two similar elementary tasks are more likely to have similar landscape modality with distinct optimal solutions than having close optima under entirely different landscape modality. %
For simplicity, we postulate that intra-family tasks share a common landscape modality but may have distinct optima in this work, which can be realized by a single-objective minimization function with configurable optimum. %
To illustrate, we present two 2-dimensional examples in Fig. \ref{fig:intrainter}. %
In the intra-family transfer, the source and target tasks with distinct optimal solutions are drawn from the Sphere family. %
In the inter-family transfer, the target task is from the Ackley family while the source tasks are sampled from the Sphere family. %
In this way, we can configure the functions and the optimal solutions of the tasks in an STOP independently. %
In the intra-family transfer, the source and target tasks have the same function. %
By contrast, the source tasks in the inter-family transfer are outer-family functions that differ from the target one. %
Now, the key task of producing a synthetic STOP is how to make its similarity distribution explicitly configurable when designing the optimal solutions of the source and target tasks. %

\subsubsection{Configurable Similarity Distribution} To enable the configurable similarity distribution, we propose to employ the inverse transform sampling~\cite{hormann2004automatic} to produce the set of similarity values $\mathbf{R^s}=\{R^s_{1},R^s_{2},...,R^s_{3}\}$ directly. Suppose the underlying similarity distribution of an STOP to be generated is denoted by $h\left(s\right)$, we can produce $k$ independent observations of similarity value from $h\left(s\right)$ as follows,
\begin{equation}
S_i=H^{-1}\left(U_i\right),\,\,\,i=1,2,...,k
\label{eq:its_generation}
\end{equation}
where $S_i$ is the $i$th observation, $H^{-1}$ denotes the generalized inverse of the cumulative distribution function $H\left(s\right)$, $U_i$ is the $i$th sample from the uniform distribution $\mathrm{Unif}[0,1]$. 

Suppose the optimum of the target task is randomly generated in the common space first and is denoted as $\mathbf{o}_{t}$, we configure the optimal solution of the $i$th source task as follows:
\begin{equation}
\mathbf{o}_{si} = \mathbf{o}_{t}+\frac{1-S_i}{\mid\mid\mathbf{r}-\mathbf{o}_{t}\mid\mid_\infty}\left(\mathbf{r}-\mathbf{o}_{t}\right)
\label{eq:optimum_source}
\end{equation}
where $\mathbf{o}_{si}$ represents the optimal solution of the $i$th source task, $\mathbf{r}\in\left[0,1\right]^d$ is a randomly generated real vector from the multivariate uniform distribution. 

There are two points worthy mentioning when using Eq. \eqref{eq:its_generation} and Eq. \eqref{eq:optimum_source}:
\begin{itemize}
\item Out-of-bounds variables in $\mathbf{o}_{si}$ will be set to their nearest bound values.
\item The minimum variable in $\mathbf{o}_{t}$ is altered to zero for enabling the similarity value of one.
\end{itemize}

According to Eq. \eqref{eq:sim_relationship}, we can calculate the similarity value between the $i$th source task and the target task as follows:
\begin{equation}
\begin{split}
R^s_i &= 1 - D_{Cheb}\left(\mathbf{o}_{si},\mathbf{o}_{t}\right)\\
&=1-\bigg|\bigg|\left(1-S_i\right)\frac{\mathbf{r}-\mathbf{o}_{t}}{\mid\mid\mathbf{r}-\mathbf{o}_{t}\mid\mid_\infty}\bigg|\bigg|_\infty\\
&=1-\left(1-S_i\right)\frac{\mid\mid\mathbf{r}-\mathbf{o}_{t}\mid\mid_\infty}{\mid\mid\mathbf{r}-\mathbf{o}_{t}\mid\mid_\infty}\\
&=S_i
\end{split}
\label{eq:sim_calculation}
\end{equation}

It can be observed that the set of similarity values $\mathbf{R^s}=\{R^s_{1},R^s_{2},...,R^s_{k}\}$ are exactly the $k$ observations from $h\left(s\right)$, i.e., $\mathbf{S}=\{S_{1},S_{2},...,S_{k}\}$. %
In other words, with Eq. \eqref{eq:its_generation} and Eq. \eqref{eq:optimum_source}, one can generate a synthetic STOP with any desired similarity relationship by explicitly specifying its underlying similarity distribution. %
This important feature is also termed as general representation of the similarity distribution, which will be theoretically analyzed in what follows.


\subsection{General Representation of the Similarity Distribution}

\begin{thm}
For any STOP with a valid latent similarity distribution $h\left(s\right)$, the proposed method can produce a synthetic STOP whose empirical similarity distribution $g\left(s;\mathbf{R}^s\right)$ converges to $h\left(s\right)$ as $k\to\infty$.
\end{thm}
\begin{IEEEproof}[Proof] For any valid similarity distribution $h\left(s\right)$ whose cumulative distribution function is denoted by $H\left(s\right)$, we can obtain $k$ independent and identically distributed observations $\mathbf{S}=[S_1,S_2,...,S_k]$ from $H\left(s\right)$ with the inverse transform sampling method, which is given by,
\begin{equation}
S=H^{-1}\left(U\right)
\notag
\end{equation}
where $H^{-1}$ denotes the generalized inverse of $H$, $U$ is uniform on $[0,1]$, i.e., $U\sim \mathrm{Unif}[0,1]$. Then, we can calculate the density estimate of $h\left(s\right)$ via the rescaled histogram as follows:
\begin{equation}
h^e\left(s;\mathbf{S}\right) = \frac{1}{k}\sum_{i=1}^{k}I_{(\frac{\lfloor sn\rfloor}{n},\frac{\lceil sn\rceil}{n}]}\left(S_i\right)
\notag
\end{equation}
where $h^e$ represents the empirical estimate of $h$, $n$ is the number of equal-width bins of the histogram. With Eq. \eqref{eq:its_generation} and Eq. \eqref{eq:optimum_source}, we have
\begin{equation}
S_i=R^s_i,\,\,\,i=1,2,...,k
\notag
\end{equation}

Then, we have
\begin{equation}
\begin{split}
g\left(s;\mathbf{R}^s\right) &= h^e\left(s;\mathbf{S}\right)\\
&=\frac{\mathrm{d}}{\mathrm{d}s}\int_{-\infty}^{s}h^e\left(w;\mathbf{S}\right)\mathrm{d}w\\
&=\frac{\mathrm{d}}{\mathrm{d}s}\int_{0}^{s}\frac{1}{k}\sum_{i=1}^{k}I_{(\frac{\lfloor wn\rfloor}{n},\frac{\lceil wn\rceil}{n}]}\left(S_i\right)\mathrm{d}w
\end{split}
\notag
\end{equation}

When $n\to\infty$, we have
\begin{equation}
\begin{split}
g\left(s;\mathbf{R}^s\right) &= \lim_{n\to\infty}\frac{\mathrm{d}}{\mathrm{d}s}\int_{0}^{s}\frac{1}{k}\sum_{i=1}^{k}I_{(\frac{\lfloor wn\rfloor}{n},\frac{\lceil wn\rceil}{n}]}\left(S_i\right)\mathrm{d}w\\
&=\frac{\mathrm{d}}{\mathrm{d}s}\frac{1}{k}\sum_{i=1}^{k}\int_{0}^{s}\lim_{n\to\infty}I_{(\frac{\lfloor wn\rfloor}{n},\frac{\lceil wn\rceil}{n}]}\left(S_i\right)\mathrm{d}w\\
&=\frac{\mathrm{d}}{\mathrm{d}s}\frac{1}{k}\sum_{i=1}^{k}\int_{0}^{s}I_w\left(S_i\right)\mathrm{d}w\\
&=\frac{\mathrm{d}}{\mathrm{d}s}\frac{1}{k}\sum_{i=1}^{k}I_{(0,s]}\left(S_i\right)
\end{split}
\notag
\end{equation}

Finally, with the strong law of large numbers, we can get
\begin{equation}
\begin{split}
g\left(s;\mathbf{R}^s\right) &=\frac{\mathrm{d}}{\mathrm{d}s}\lim_{k\to\infty}\frac{1}{k}\sum_{i=1}^{k}I_{(0,s]}\left(S_i\right)\\
&=\frac{\mathrm{d}}{\mathrm{d}s}\mathrm{Pr}\lbrack-\infty<w\le s\rbrack\\
&=\frac{\mathrm{d}}{\mathrm{d}s}\int_{-\infty}^{s}h\left(w\right)\mathrm{d}w\\
&=\frac{\mathrm{d}}{\mathrm{d}s}H\left(s\right)\\
&=h\left(s\right)
\end{split}
\notag
\end{equation}

\end{IEEEproof}

\subsection{A Problem Generator}

According to the design principles discussed above, one can construct an STOP by instantiating two main parameters: the transfer scenario and the similarity distribution. %
Algorithm 1 provides the detailed implementation of the proposed problem generator. %
When generating the source and target tasks, we configure their functions and optimal solutions independently. %
Firstly, the similarity values allocated to the $k$ source tasks are sampled from the given similarity distribution using the inverse transform sampling method in Eq. \eqref{eq:its_generation}. %
After that, the optimum of the target task is randomly sampled within the common space, which is followed by the construction of the target task based on the specified family in a set of candidate families. %
Subsequently, the optimal solutions of the $k$ source tasks are determined based on the similarity values and the target optimum according to Eq. \eqref{eq:optimum_source}, followed by the construction of the source tasks based on the specified transfer scenario, as shown in lines 5 to 10. %
For brevity, the intra-family and inter-family transfers are represented by $T_a$ and $T_e$, respectively. %
Lastly, we need to optimize the $k$ source tasks using the specified backbone optimizer and store their searching-related data into the knowledge base $\mathcal{M}$. %
It should be noted that the backbone optimizer here can be any of trial-and-error searching algorithms including, but not limited to, coordinate descent methods, evolutionary algorithms, and machine learning-assisted searching algorithms, etc. %
This optimizer-independent feature enables the problem generator coherent with the state-of-the-art in optimizers. %
Following the implementation in Algorithm 1, we name an STOP as $\mathcal{F}$-$\mathcal{T}$-$h$-$d$-$k$, where $\mathcal{F}$ denotes the target family, $\mathcal{T}$ represents the transfer scenario, $h$ is the similarity distribution, $d$ denotes the task dimension, $k$ is the number of source tasks.

\begin{algorithm}
\begin{small}
\caption{A Problem Generator of STOP}
\KwIn{$\Upsilon$ (candidate families), $i_t$ (the index of the target family), $\mathcal{T}$ (the transfer scenario), $h\left(s\right)$ (the similarity distribution), $d$ (the task dimension), $k$ (the number of source tasks), $\mathbb{O}$ (the backbone optimizer)}
\KwOut{$f\left(\mathbf{x}\right)$ (the target task), $\mathcal{M}$ (the knowledge base)}
\tcp{Generate the source and target tasks}
$\mathbf{S}\leftarrow$ Eq. \eqref{eq:its_generation} with the given $h\left(s\right)$\;
$\mathbf{o}_t\leftarrow$A randomly generated vector in the common space\;
$f\left(\mathbf{x}\right)\leftarrow\Upsilon\left(i_t,\mathbf{o}_t\right)$\;
$\left[\mathbf{o}_t,\mathbf{o}_{s1},...,\mathbf{o}_{sk}\right]\leftarrow$ Eq. \eqref{eq:optimum_source} using $\mathbf{o}_t$ and $\mathbf{S}$\;
\For{$i=1\,\,\mathrm{to}\,\,k$}
{
	\If(\tcp*[h]{intra-family transfer}){$\mathcal{T}=T_a$}
	{
		$i_s\leftarrow i_t$\;
	}
	\Else(\tcp*[h]{inter-family transfer})
	{
		$i_s\leftarrow$Random selection in $i_\Upsilon\setminus i_t$\;
	}
	$f^s_i\left(\mathbf{x}\right)\leftarrow\Upsilon\left(i_s,\mathbf{o}_{si}\right)$\;
}
\tcp{Construct the knowledge base}
$\mathcal{M}=\emptyset$\;
\For{$i=1\,\,\mathrm{to}\,\,k$}
{
	$D^{si} = \mathrm{Optimization}\left(\mathbb{O},f^s_i\right)$\;
	$\mathcal{M}\leftarrow\mathcal{M}\cup D^{si}$\;
}
\end{small}
\end{algorithm}

\begin{table}[ht]
	\caption{Important features of our test problems against those in the existing studies.}
	\centering
	\footnotesize
	\heavyrulewidth=0.12em
	\lightrulewidth=0.1em
	\cmidrulewidth=0.1em
	\setlength\tabcolsep{9pt}
	\begin{tabular}{*{4}{llllll}}
		\toprule
		Publications&\makecell[l]{Transfer\\Scenario}&\makecell[l]{Similarity\\Distribution}&\makecell[l]{Number of\\Source Tasks}\\
		\midrule
        \cite{da2018curbing,zhou2021learnable,xue2021evolutionary}&\makecell[l]{intra-family\\ \& inter-family}&fixed&fixed\\
		\midrule
		\cite{jiang2023block,wu2023transferable,liu2023multifactorial}&inter-family&fixed&fixed\\
		\midrule
         \cite{louis2004learning,zhang2019multisource}&intra-family&fixed&adjustable\\
		\midrule
        Ours&\makecell[l]{intra-family\\ \& inter-family}&customizable&adjustable\\
		\bottomrule
	\end{tabular}
	\label{tab:problem_comparison}
\end{table}

\subsection{Comparison with the Existing Test Problems}

Table \ref{tab:problem_comparison} compares the important features of our test problems against those used in the preceding studies. %
In \cite{feng2017autoencoding,da2018curbing,zhou2021learnable}, the problems are made by the commonly used benchmark functions in multiobjective optimization. %
The source and target tasks in these STOPs have distinct objective functions but possess the same optima, which can be seen as a specific class of problems produced by the generator with $\mathcal{T}=T_e$ and $h$ of the Dirac delta distribution at $s=1$. %
These problems lacks the diversity of the similarity relationship. %
To mimic the heterogeneity in terms of optimum, the authors in \cite{xue2021evolutionary} use the randomly generated optima to configure their STOPs. %
However, the similarity distribution corresponding to such randomly generated optima cannot be explicitly specified in advance and stills lacks the desired diversity. %
Moreover, the number of source tasks in all the above STOPs is not configurable. %
In \cite{louis2004learning,zhang2019multisource}, the source tasks are synthesized based on the target task. %
However, the diversity of the generated source tasks is greatly constrained to ensure the source-target similarity, leading to a specific similarity distribution with incredibly high similarity values. %
In this sense, the problems in \cite{louis2004learning,zhang2019multisource} are a particular subset of our test problems, which can be produced by the generator using a specific similarity distribution with high similarity values. %

In summary, the highlights of our proposed problem generator are as follows. %
(1) The number of source tasks is adjustable. %
(2) Both the intra- and inter-family transfer scenarios are available. %
(3) The customizable similarity distribution enables a continuous spectrum of representation of the similarity relationship between the optima of the source and target tasks, leading to a closer resemblance of the diverse similarity relationship in real-world problems.

\section{Benchmark Problems}
In this section, we design a benchmark suite containing 12 STOPs. %
Firstly, the available realizations of the parameters for building STOPs are provided. %
Then, we develop a benchmark suite with detailed problem specifications. %
Lastly, the evaluation criterion for comparing different algorithms on the developed benchmark suite is provided.

\subsection{Available Realizations of the Parameters}

According to Algorithm 1, we can see that there are five parameters needed for building STOPs: the task family, the transfer scenario, the similarity distribution, the number of source tasks, and the task dimension. %
In what follows, the available realizations of these five parameters are presented. %

\subsubsection{Task Families} We choose eight single-objective functions with configurable optimum to serve as the candidate families due to their widespread use in the area of continuous optimization.

-- Sphere family:
\begin{equation}
\notag
\small
\min f_1\left(\mathbf{x}\right)=\sum_{i=1}^d\left(x_i-o_i\right)^2,\,\mathbf{x}\in\left[-100,100\right]
\label{eq:sphere}
\end{equation}

-- Ellipsoid family:
\begin{equation}
\notag
\small
\min f_2\left(\mathbf{x}\right)=\sum_{i=1}^d\left(d-i+1\right)\left(x_i-o_i\right)^2,\,\mathbf{x}\in\left[-50,50\right]
\label{eq:ellipsoid}
\end{equation}

-- Schwefel 2.2 family:
\begin{equation}
\notag
\small
\min f_3\left(\mathbf{x}\right)=\sum_{i=1}^d|x_i-o_i|+\prod_{i=1}^d|x_i-o_i|,\,\mathbf{x}\in\left[-30,30\right]
\label{eq:schwefel}
\end{equation}

-- Quartic family with noise:
\begin{equation}
\notag
\small
\min f_4\left(\mathbf{x}\right)=\varepsilon+\sum_{i=1}^d i\times\left(x_i-o_i\right)^4,\,\mathbf{x}\in\left[-5,5\right]
\label{eq:quartic}
\end{equation}

-- Ackley family:
\begin{equation}
\notag
\footnotesize
\begin{split}
\min f_5\left(\mathbf{x}\right)=&-20\mathrm{exp}\left(-0.2\sqrt{\frac{1}{D}\sum_{i=1}^{D}z_i^2}\right)-\mathrm{exp}\left(\frac{1}{D}\sum_{i=1}^{D}\mathrm{cos}\left(2\pi z_i\right)\right)\\
&+20+e,\,z_i=x_i-o_i,\,\mathbf{x}\in\left[-32,32\right]
\end{split}
\label{eq:ackley}
\end{equation}

-- Rastrigin family:
\begin{equation}
\notag
\small
\begin{split}
\min f_6\left(\mathbf{x}\right)=&\sum_{i=1}^d\Big\{\left(x_i-o_i\right)^2+10\mathrm{cos}\big\lbrack2\pi\left(x_i-o_i\right)\big\rbrack\Big\}\\
&+10d,\,\mathbf{x}\in\left[-10,10\right]
\end{split}
\label{eq:rastrigin}
\end{equation}

\begin{figure}[ht]
	\centering
	\includegraphics[width=2.7in]{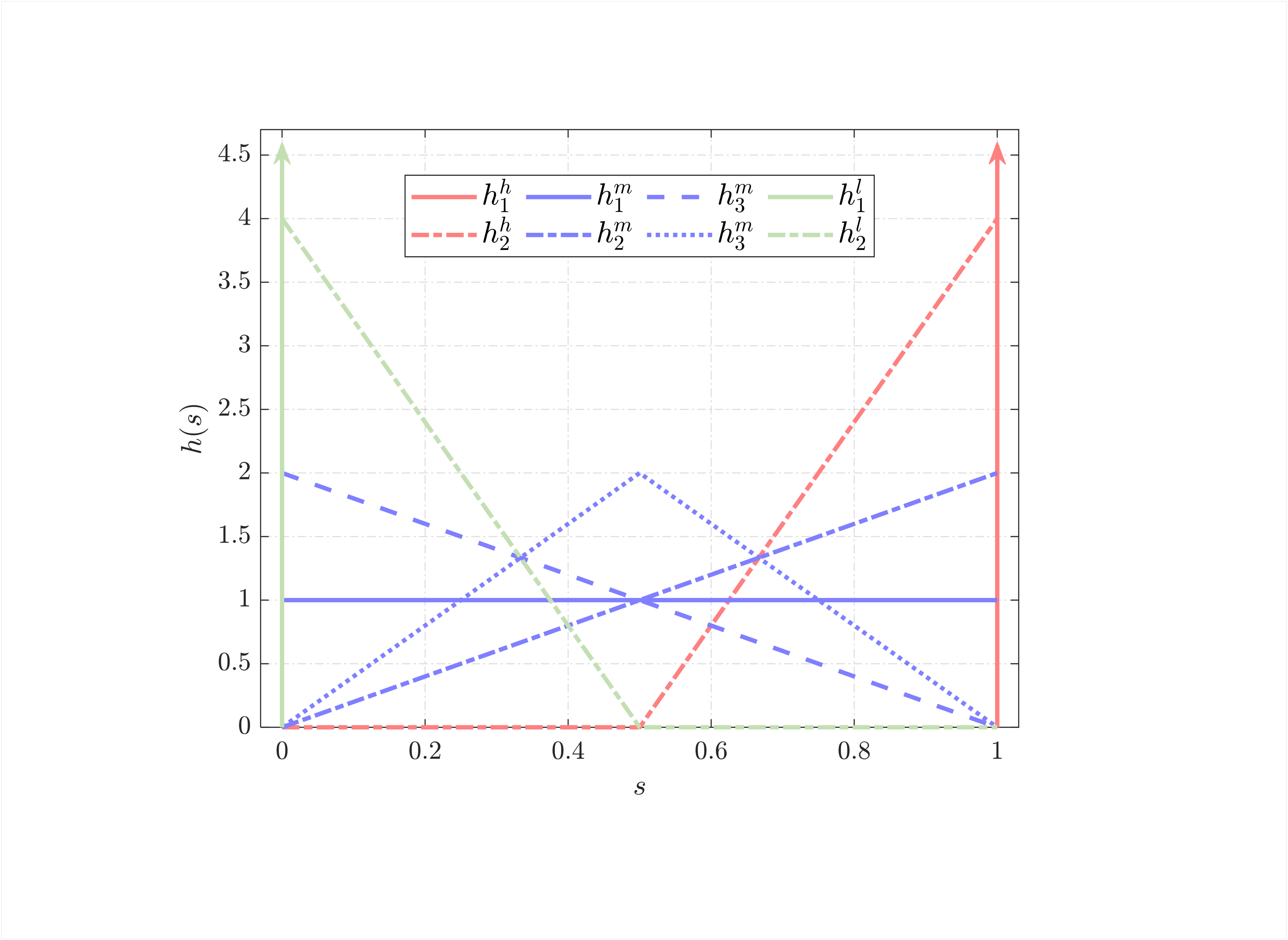}
	\caption{Illustration of the eight similarity distributions.}
	\label{fig:sim_benchmarks}
\end{figure}

-- Griewank family:
\begin{equation}
\notag
\small
\begin{split}
\min f_7\left(\mathbf{x}\right)=&1+\frac{1}{4000}\sum_{i=1}^{d}\left(x_i-o_i\right)^2\\
&-\prod_{i=1}^d\mathrm{cos}\left(\frac{x_i-o_i}{\sqrt{i}}\right),\,\mathbf{x}\in\left[-200,200\right]
\end{split}
\label{eq:griewank}
\end{equation}

-- Levy family:
\begin{equation}
\notag
\footnotesize
\begin{split}
\min f_8\left(\mathbf{x}\right)=&\mathrm{sin}^2\left(\pi\omega_1\right)+\sum_{i=1}^{d-1}\left(\omega_i-1\right)^2\left[1+10\mathrm{sin}^2\left(\pi\omega_i+1\right)\right]+\\
&\left(\omega_{d}-1\right)^2\left[1+\mathrm{sin}^2\left(2\pi\omega_{d}\right)\right],\,\omega_i=1+\frac{z_i}{4},\,\mathbf{x}\in\left[-20,20\right]
\end{split}
\label{eq:levy}
\end{equation}
where $d$ is the dimension, $x_i$ denotes the $i$th decision variable, and $o_i$ represents the optimal solution of the $i$th variable, $\varepsilon\sim U\left(0,1\right)$ denotes the random noise. %
The eight functions here act as the set of candidate families (i.e., $\Upsilon$) in Algorithm 1. %
It should be noted that the design here is not limited to the eight selected functions and can be generalized to any set of functions with configurable optimum.

\subsubsection{Transfer Scenarios} Both the intra-family and inter-family transfers are considered in this work. %
In the intra-family transfer, the source and target tasks of an STOP are from the same family. %
By contrast, the source tasks in the inter-family transfer are from the families that differ from the target one.

\subsubsection{Similarity Distributions} Three types of similarity distributions with different levels of similarity values are considered in this study, which are marked by different colors and shown in Fig. \ref{fig:sim_benchmarks}. %
The two similarity distributions with high similarity values are presented as follows:

-- High similarity 1:
\begin{equation}
\begin{split}
h_1^h\left(s\right)&=\delta\left(s-1\right)\\
&=\lim _{\sigma \rightarrow 0} \frac{1}{|\sigma| \sqrt{\pi}} e^{-[(s-1)/ \sigma]^2}
\end{split}
\notag
\end{equation}

-- High similarity 2:
\begin{equation}
\begin{split}
h_2^h\left(s\right)&=\mathrm{ReLU}\left(8s-4\right)\\
&=\frac{\left(8s-4\right)+\mid8s-4\mid}{2}
\end{split}
\notag
\end{equation}

The four similarity distributions with mixed similarity values are as follows:

\begin{table}[ht]
	\caption{Available realizations of the five parameters.}
	\centering
	\heavyrulewidth=0.12em
	\lightrulewidth=0.1em
	\cmidrulewidth=0.1em
	\setlength\tabcolsep{6pt}
	\begin{tabular}{*{2}{ll}}
		\toprule
		Parameter&Configuration\\
		\midrule
		The task family ($\Upsilon$)&$\{f_1, f_2, f_3, f_4, f_5, f_6, f_7, f_8\}$\\
		The transfer scenario ($\mathcal{T}$)&$\{T_a, T_e\}$\\
		The similarity distribution ($h$)&$\{h^h_1,h^h_2,h^m_1,h^m_2,h^m_3,h^m_4,h^l_1,h^l_2\}$\\
		The task dimension ($d$)&$N+$\\
		The number of source tasks ($k$)&$N+$\\
		\bottomrule
	\end{tabular}
	\label{tab:summary_aspects}
\end{table}

-- Mixed similarity 1:
\begin{equation}
h_1^m\left(s\right)=1
\notag
\end{equation}

-- Mixed similarity 2:
\begin{equation}
h_2^m\left(s\right)=2s
\notag
\end{equation}

-- Mixed similarity 3:
\begin{equation}
h_3^m\left(s\right)=2-2s
\notag
\end{equation}

-- Mixed similarity 4:
\begin{equation}
h_4^m\left(s\right)=
\begin{cases}
4s,\,\,\,0\le s\le 0.5\\
4-4s,\,\,\,0.5\le s\le 1\\
\end{cases}
\notag
\end{equation}

The two similarity distributions with low similarity values are as follows:

-- Low similarity 1:
\begin{equation}
\begin{split}
h_1^l\left(s\right)&=\delta\left(s\right)\\
&=\lim _{\sigma \rightarrow 0} \frac{1}{|\sigma| \sqrt{\pi}} e^{-(s / \sigma)^2}
\end{split}
\notag
\end{equation}

-- Low similarity 2:
\begin{equation}
\begin{split}
h_2^l\left(s\right)&=\mathrm{ReLU}\left(4-8s\right)\\
&=\frac{\left(4-8s\right)+\mid4-8s\mid}{2}
\end{split}
\notag
\end{equation}

\subsubsection{Task Dimensions} Without loss of generality, we suggest the problem dimensions from 25 to 50 to avoid the curse of dimensionality. %
It should be noted that the dimensional heterogeneity is not considered herein. %
Thus, different STOPs can be configured with different dimensions, but the source and target tasks in a single STOP have the same dimensions.

\subsubsection{The Number of Source Tasks} The valid numbers of source tasks are positive integers. %
Both the single-source and multi-source scenarios can be set up using the proposed generator. %
Table \ref{tab:summary_aspects} summarizes the available realizations of the five parameters. %
Next, we shall formulate a benchmark suite with 12 individual STOPs.

\subsection{A Benchmark Suite}

\begin{table}[ht]
	\caption{A benchmark suite of STOPs.}
	\centering
	\heavyrulewidth=0.12em
	\lightrulewidth=0.1em
	\cmidrulewidth=0.1em
	\setlength\tabcolsep{9pt}
	\begin{tabular}{*{3}{lll}}
		\toprule
		\makecell[l]{Similarity\\Relationship}&\makecell[l]{Problem Specification\\ ($\mathcal{F}$-$\mathcal{T}$-$h$-$d$-$k$)} &Problem ID\\
		\midrule
		\multirow{7}{*}{\makecell[l]{High Similarity (HS)}}&Sphere-$T_a$-$h^h_1$-50-$k$&STOP 1\\
		\cmidrule(l){2-3}
		&Ellipsoid-$T_e$-$h^h_2$-25-$k$&STOP 2\\
		\cmidrule(l){2-3}
		&Schwefel-$T_a$-$h^h_2$-30-$k$&STOP 3\\
		\cmidrule(l){2-3}
		&Quartic-$T_e$-$h^h_1$-50-$k$&STOP 4\\
		\midrule
		\multirow{7}{*}{\makecell[l]{Mixed Similarity (MS)}}&Ackley-$T_a$-$h^m_1$-25-$k$&STOP 5\\
		\cmidrule(l){2-3}
		&Rastrigin-$T_e$-$h^m_2$-50-$k$&STOP 6\\
		\cmidrule(l){2-3}
		&Griewank-$T_a$-$h^m_3$-25-$k$&STOP 7\\
		\cmidrule(l){2-3}
		&Levy-$T_e$-$h^m_4$-30-$k$&STOP 8\\
		\midrule
		\multirow{7}{*}{\makecell[l]{Low Similarity (LS)}}&Sphere-$T_a$-$h^l_1$-25-$k$&STOP 9\\
		\cmidrule(l){2-3}
		&Rastrigin-$T_e$-$h^l_2$-30-$k$&STOP 10\\
		\cmidrule(l){2-3}
		&Ackley-$T_a$-$h^l_2$-50-$k$&STOP 11\\
		\cmidrule(l){2-3}
		&Ellipsoid-$T_e$-$h^l_1$-50-$k$&STOP 12\\
		\bottomrule
	\end{tabular}
	\label{tab:benchmark_suite}
\end{table}

One of key rationales behind designing benchmark problems is to provide a set of representative test problems with diverse properties, which can create a good resemblance to a wide range of real-world problems~\cite{omidvar2015designing}. %
Thus, the configurations in Table \ref{tab:summary_aspects} should be adequately considered when designing a benchmark suite of STOPs. %
According to the three types of similarity distributions in Fig. \ref{fig:sim_benchmarks}, we develop three categories of STOPs: 1) STOPs full of source tasks that are highly similar to the target task in terms of optimum; 2) STOPs containing source tasks with mixed similarity to the target task; 3) STOPs full of source tasks with low similarity to the target task. %
These three categories of STOPs are termed HS, MS, and LS problems for short, respectively, each of which contains four individual problems. %
The eight task families and the two transfer scenarios are alternately configured across the 12 problems. %
The number of source tasks is a user-defined parameter, which can be customized to cater to the needs of studies on transfer with different amount of available optimization experience.

Table \ref{tab:benchmark_suite} lists the twelve problems in the benchmark suite, which can be readily produced by the proposed generator according to the specifications. %
For the first category of problems from STOP 1 to STOP 4, the optimal solutions of the source and target tasks are close in the common space, making the optimal solutions of all the source tasks are highly transferable for accelerating the target search. %
For the second category of problems, the four customized similarity distributions (i.e., $h^m_1$ to $h^m_4$) are used for adjusting the proportion of different levels of similarity values. %
As for STOP 9 to STOP 12, the optimal solutions of the source tasks are dissimilar to the target one. %
Consequently, the optimal solutions of all the source tasks in STOP 9 to STOP 12 are always equally unhelpful for speeding up the target search. %
It is worth mentioning that the similarity here indicates that whether the unadapted optimal solution of a source task is similar to the target one. %
If the solution can be adapted properly, its similarity to the target optimum can be improved a lot.

\subsection{Evaluation and Comparative Results} The minimum value of all the target tasks in the benchmark suite is zero. %
When comparing multiple algorithms, the computational budget in terms of function evaluation allocated to each algorithm should be identical. %
To examine the significance of numerical results, we recommend comparison methods that use the statistical significance tests in \cite{derrac2011practical}.

\begin{figure}[ht]
	\centering
	\includegraphics[width=3.4in]{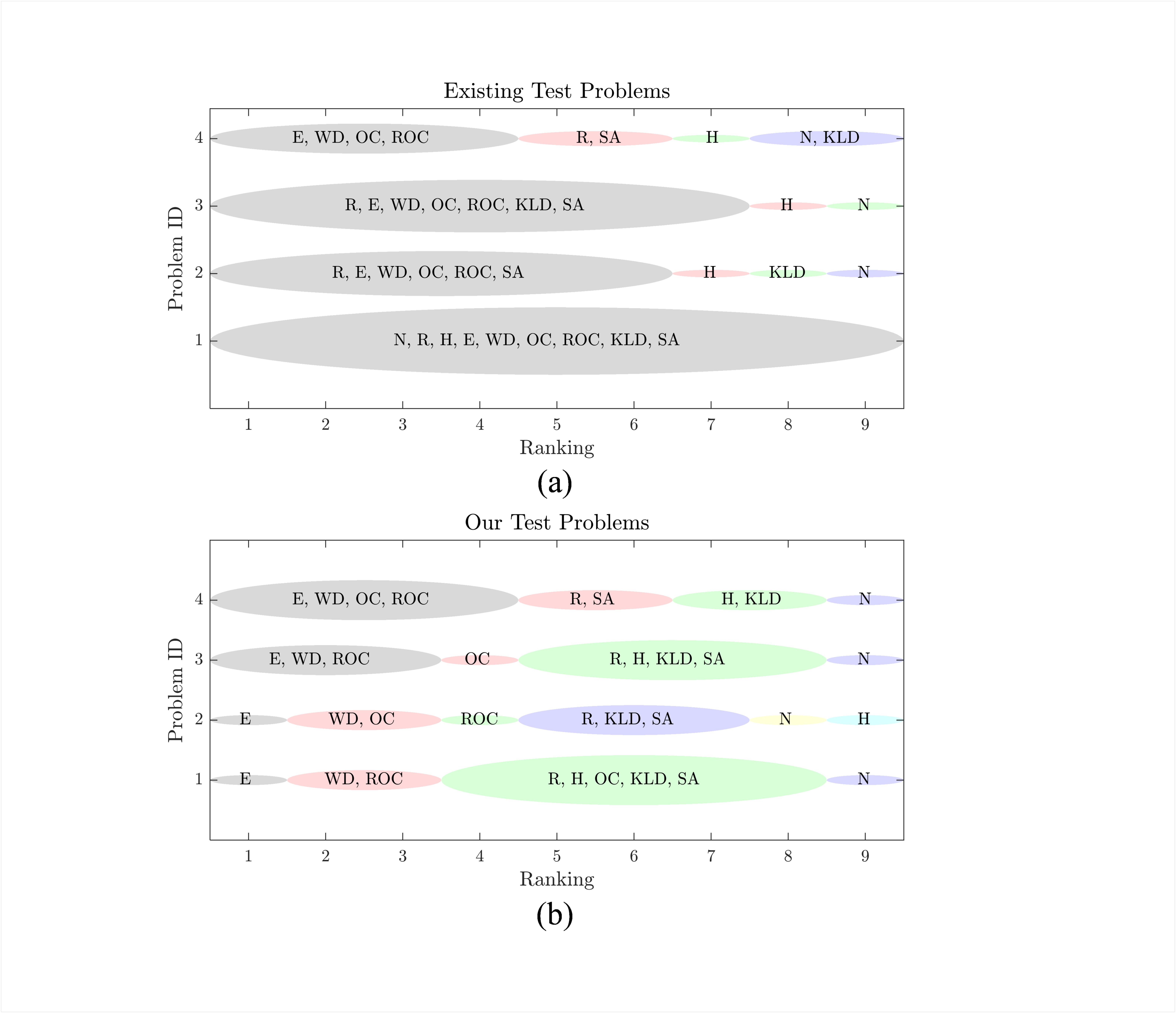}
	\caption{Performance rankings of the two baseline and seven source selection-based STO algorithms: (a) the existing test problems; (b) our test problems.}
	\label{fig:rankings}
\end{figure}

To empirically demonstrate the superiority of our test problems against those in preceding studies, we conduct a comparative study with a number of source selection-based STO algorithms. %
Following the synthesis of test problems in existing studies, we form four STOPs standing for existing test problems with the ingredient functions in~\cite{Da2016Evolutionary}. %
For each STOP whose target task is specified as a particular function, the remaining ingredient functions act as the source tasks. %
Two baselines with no transfer and random source selection, as denoted by N and R, and seven source selection methods based on similarity measurement are considered here, including the hamming distance (H)~\cite{louis2004learning}, Euclidean distance (E)~\cite{ding2019generalized}, Kullback-Leibler divergence (KLD)~\cite{huang2021surrogate}, Wasserstein distance (WD)~\cite{zhang2019multisource}, ordinal correlation (OC)~\cite{zhou2021learnable}, relaxed ordinal correlation (ROC)~\cite{xue2021evolutionary}, and subspace alignment (SA)~\cite{tang2020regularized}. %
The backbone optimizer used in all the STO algorithms is evolutionary algorithm with the simulated binary crossover, polynomial mutation and 1/2 truncation selection, in which $p_c$, $\eta_c$, $p_m$, $\eta_m$ are set to be 1, 15, $1/d$ and 15, respectively. %
The population size is set to 50 while the maximum number of functions evaluations available is 5$\times 10^3$. %
The number of independent runs for each algorithm is 30. %

Fig. \ref{fig:rankings} shows the rankings of the algorithms on the four existing test problems and the four test problems with the mixed similarity developed in this study. %
The rankings are determined by the Wilcoxon rank-sum test on the final objective values over 30 independent runs. %
Any two algorithms without the statistical significance in terms of their final objective values will be put into the same shaded group. %
From Fig. \ref{fig:rankings}(a), we can see that most of the algorithms show tied rankings across the existing test problems, indicating the problems' incapability of assessing the source selection methods. %
This phenomenon is largely attributed to the limited patterns of similarity relationships between the source and target tasks of the four problems. %
The similarity values of the source tasks to the target task are either very high or very low, making most of the source selection-based STO algorithms either achieve indistinguishably better results than N or fail to speed up the target search simultaneously, as shown in Fig. \ref{fig:rankings}(a). %
By contrast, different levels of rankings of the algorithms can be observed from Fig. \ref{fig:rankings}(b). %
The results here indicate that the proposed STOPs with the mixed similarity can serve as an arena for helping researchers identify and design effective source selection methods. %
In summary, with the broad spectrum of representation of the most pertinent feature of STOPs, i.e., similarity distribution, our test problems are able to provide a more comprehensive investigation of the performance of STO algorithms as compared to the existing test problems. %


\section{Conclusion}

In this study, we have proposed several basic concepts to characterize the tasks in an STOP, including the task family, task-optimum mapping, optimum coverage, and similarity distribution. %
In particular, the similarity distribution is an important problem feature of STOPs, as the knowledge reasoning in most transfer optimization algorithms is basically analogy: the more similar two tasks are, the more likely the search experience is to be transferred between them towards enhanced optimization performance. %
This problem feature always plays a significant role in the performance of STO algorithms, especially those that transfer elite solutions or model information based upon solution data. %
Thus, the similarity distribution should be adequately considered when designing benchmarks for STOPs. %

Since the similarity distribution is problem-dependent, we have proposed to develop a scalable test problem generator that is able to generate STOPs with customizable similarity distributions. %
Specifically, with a novel optimum configuration scheme based on the inverse transform sampling, one can explicitly specify the desired similarity distribution for problem generation. %
It is noted that the proposed problem generator shows superior extendability and scalability, enabling it to generate STOPs that cater to the unique needs of individual studies. %
To the best of our knowledge, this study serves as the first work regarding the systematic generation of synthetic black-box STOPs. %

In future work, more problem complexities would be taken into account when designing synthetic STOPs, including objective conflicts in multi-objective optimization, high-dimensionality in large-scale optimization, and constraint satisfaction in constrained problems.

\ifCLASSOPTIONcaptionsoff
  \newpage
\fi



%

\footnotesize
\bibliography{MyBiBfile}

%





\end{document}